\documentclass{article}

\usepackage{microtype}
\usepackage{graphicx}
\usepackage{subfigure}
\usepackage{booktabs} 
\usepackage{amssymb}
\usepackage{tikz}
\usepackage{soul}
\usepackage[noend]{algpseudocode}
\usepackage{makecell}
\usepackage{multirow}
\usepackage{multicol}
\usepackage{enumitem}

\setitemize{topsep=0pt, itemsep=0pt, parsep=2pt}
\setenumerate{topsep=0pt, itemsep=0pt, parsep=2pt}

\newcommand{\correction}{}

\usepackage{hyperref}

\usepackage[accepted]{mlsys2023}

\makeatletter
\renewcommand{\mlsys@appearing}{Pre-print. Do not redistribute.}
\makeatother

\mlsystitlerunning{Pex: Memory-efficient Microcontroller Deep Learning through Partial Execution}

\begin{document}

\newcommand\opexec{\stackrel{op}{\longrightarrow}}
\newcommand{\boxlabel}[1]{\tikz[baseline=(X.base)]\node [draw=black,fill=white,semithick,rectangle,inner sep=1.5pt] (X) {#1};}

\twocolumn[
\mlsystitle{Pex: Memory-efficient Microcontroller \\ Deep Learning through Partial Execution}



\mlsyssetsymbol{equal}{*}

\begin{mlsysauthorlist}
\mlsysauthor{Edgar Liberis}{cam,sam}
\mlsysauthor{Nicholas D. Lane}{cam,sam}
\end{mlsysauthorlist}

\mlsysaffiliation{cam}{Department of Computer Science and Technology, University of Cambridge, UK}
\mlsysaffiliation{sam}{Samsung AI Centre Cambridge, UK}

\mlsyscorrespondingauthor{Edgar Liberis}{el398 \textit{at} cam.ac.uk}

\mlsyskeywords{Machine Learning, MLSys, MCU, NAS, Deep learning, Neural Networks, Model Compression, Partial execution, Microcontrollers, TinyML}

\vskip 0.3in

\begin{abstract}
Embedded and IoT devices, largely powered by microcontroller units (MCUs), could be made more intelligent by leveraging on-device deep learning. One of the main challenges of neural network inference on an MCU is the extremely limited amount of read-write on-chip memory (SRAM, $<$ 512 kB). SRAM is consumed by the neural network layer (operator) input and output buffers, which, traditionally, must be in memory (materialised) for an operator to execute. 
We discuss a novel execution paradigm for microcontroller deep learning, which modifies the execution of neural networks to avoid materialising full buffers in memory, drastically reducing SRAM usage with no computation overhead. 
This is achieved by exploiting the properties of operators, which can consume/produce a fraction of their input/output at a time.
We describe a \underline{p}artial \underline{ex}ecution compiler, \textsc{Pex}, which produces memory-efficient execution schedules automatically by identifying subgraphs of operators whose execution can be split along the feature/``channel'' dimension.
Memory usage is reduced further by targeting memory bottlenecks with structured pruning, leading to the co-design of the network architecture and its execution schedule. Our evaluation of image and audio classification models: (a) establishes state-of-the-art performance in low SRAM usage regimes for considered tasks with up to +2.9\% accuracy increase; (b) finds that a 4$\times$ memory reduction is possible by applying partial execution alone, or up to 10.5$\times$ when using the compiler-pruning co-design, while maintaining the classification accuracy compared to prior work; (c) uses the recovered SRAM to process higher resolution inputs instead, increasing accuracy by up to +3.9\% on Visual Wake Words.
\end{abstract}
]



\printAffiliationsAndNotice{}  

\section{Introduction}
\label{sec:introduction}

The low cost and versatility of microcontroller platforms have made them an attractive choice for a wide range of applications: an estimated 29B units shipped in 2021~\cite{mcumarket2021}. This includes numerous embedded, personal and IoT devices, many of which could be enhanced with computational intelligence brought by deep learning.

A microcontroller unit (MCU) is a single chip mainly consisting of a power-efficient CPU, read-only Flash memory and read-write static RAM (SRAM). Table~\ref{tab:devices-comp} compares MCUs along compute, memory and storage capacity axes with GPU server and mobile hardware capable of executing neural networks. The low price and power usage of MCUs come with a marked downgrade in the computational ability: memory capacity lies orders of magnitude behind the next-best hardware, and there is little-to-no memory hierarchy and parallelism to exploit for optimisation. This presents a significant challenge to deep learning inference on the device itself, causing applications to rely on fully remote or hybrid deployments~\cite{almeida2021dyno}, which results in sacrifices in data privacy and autonomy.

\begin{table}
\caption{Specifications of a high-end GPU, a smartphone, a micro-PC and Nucleo development boards using Cortex M7 and M4 chips. MCUs have the most favourable power efficiency and price but are significantly resource-constrained.}

\label{tab:devices-comp}
\centering
\footnotesize
\begin{tabular}{rrrrr}
\toprule
\multicolumn{1}{l}{\textbf{CPU}} & \multicolumn{1}{l}{\textbf{RAM}} & \multicolumn{1}{l}{\textbf{Storage}} & \multicolumn{1}{l}{\textbf{Power}} & \multicolumn{1}{l}{\textbf{Price}} \\
\midrule
\multicolumn{5}{l}{\textbf{GPU:} NVIDIA A100 GPU} \\
6912$\times$@1.4 GHz & 40 GB & --- & 400 W & 12.5K\$ \\
\midrule
\multicolumn{5}{l}{\textbf{Mobile:} Galaxy S22 Smartphone} \\
8$\times$ @ $<$2.8 GHz & 8 GB & 128 GB & $\approx$3.6 W & 750\$ \\
\midrule
\multicolumn{5}{l}{\textbf{Micro-PC:} Raspberry Pi 4B} \\
4$\times$ @1.5 GHz & $<$4 GB & 512 GB & 2.7 W & 35\$ \\
\midrule
\multicolumn{5}{l}{\textbf{MCU:} \href{https://os.mbed.com/platforms/ST-Nucleo-F767ZI/}{NUCLEO-F767ZI}~(M7)~\cite{nucleof767zi}} \\
216 MHz & 512 KB & 2 MB & 0.3 W & 9\$ \\
\midrule
\multicolumn{5}{l}{\textbf{MCU:} \href{https://os.mbed.com/platforms/ST-Nucleo-F446RE/}{NUCLEO-F446RE}~(M4)~\cite{nucleof446re}} \\
180 MHz & 128 KB & 512 KB & 0.1 W & 3\$ \\
\bottomrule
\end{tabular}
\end{table}

\begin{figure*}[ht!]
    \centering
    \includegraphics[width=\textwidth]{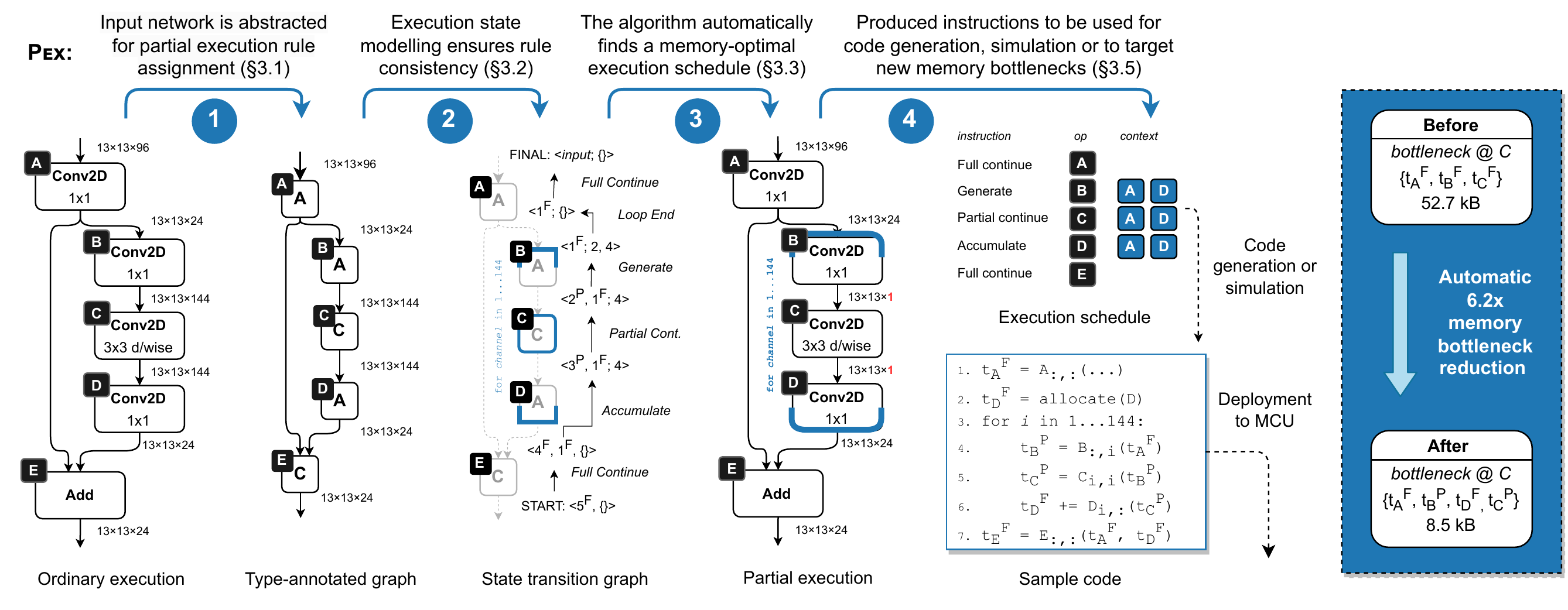}%
    \caption{\correction \textsc{Pex} compilation illustrated using the MobileNet-v2 IRB example. \textsc{Pex} operates on the network computation graph and assigns the devised partial execution rules (Section~\ref{sec:pexec-rules}) to each operator. The rules determine whether an operator will produce/consume full or partial (1 channel) input/output, enabling a drastic reduction in memory usage when partial computation is possible. To ensure an inter-consistent and automatic assignment, \textsc{Pex} models the memory state during execution (Section~\ref{sec:execution-state}) and leverages a dynamic programming-based scheduler (Section~\ref{sec:pexec-algorithm}). The result is the instruction list---a topologically ordered list of operators and assigned execution rules, and other contextual information---which is sufficient to produce native code for subsequent microcontroller deployment.}
    \label{fig:pex}
\end{figure*}

Enabling on-device network inference has been shown to require specialised research into low-footprint deep learning. TinyML~\cite{warden2020tinyml} is a research direction that encompasses hardware improvements~\cite{sadiq2022tinyops}; systems-level software research, such as runtime~\cite{david2020tensorflow}, layer implementations~\cite{lai2018cmsis} and model compilers~\cite{chen2018tvm}; model discovery,  compression and architecture search (NAS)~\cite{liberis2021munas, banbury2020micronets, lin2020mcunet}.

\subsection{Memory usage of neural networks on MCUs}
In this work, we focus on the issue of high memory (SRAM) usage of neural network inference, which prevents the MCU deployment of models with large activation tensors. 

A neural network is defined as a computational graph of layers, called \emph{operators}, with data dependencies between them. Most MCU deep learning runtimes and kernel implementations~\cite{chen2018tvm, david2020tensorflow, lai2018cmsis}, have adopted a \emph{one-operator-at-a-time} execution regime. For each layer/operator in the network, the runtime: 
\begin{enumerate}
    \item Allocates the output buffer for the operator in SRAM; 
    \item Executes the operator, using inputs read from SRAM and parameters read from the separate Flash storage; 
    \item Marks the operator’s inputs’ buffers as available memory (\emph{i.e.} deallocates from SRAM, may be implicit), if the buffers are not used later.
\end{enumerate}

Under this execution regime, called \emph{ordinary execution}, the memory bottleneck is the size of the largest operator's working set (input and output buffers) along with any other tensors that must be retained for subsequent operators. This bottleneck must lie within the SRAM capacity constraints. 

Some systems-level model-agnostic methods have been developed to reduce SRAM usage of neural networks by optimising buffer layout in memory, such as buffer bin-packing~\cite{david2020tensorflow} and operator reordering~\cite{liberis2019neural}. SRAM usage is also an optimisation objective in MCU-compatible model discovery methods~\cite{fedorov2019sparse, liberis2021differentiable, banbury2020micronets}. 

\subsection{\textsc{Pex}: Partial execution for memory efficiency}
A core contribution of this paper is a novel model compiler, called \textsc{Pex}, which realises further gains in memory usage by moving beyond the paradigm of executing one operator at a time and instead executing operators \emph{partially} to produce and/or consume \emph{one feature/channel of data at a time}. The idea of \emph{partial execution} can be applied to arbitrary neural networks. A particular instance of this was first described in MobileNet-v2~\cite{sandler2018mobilenetv2} as a memory-efficient trick to execute the inverted residual block (IRB). Figure~\ref{fig:pex} uses this example to illustrate the concept of partial execution, both diagrammatically and in pseudo-code.

Under ordinary execution, the IRB memory bottleneck resides at the operator \boxlabel{C}: the bottleneck consists of the input and output buffers of \boxlabel{C}, as well as output of \boxlabel{A} retained for operator \boxlabel{E}, amounting to 52.7 KB (at 8-bit precision).

However, operators \boxlabel{B}, \boxlabel{C}, and \boxlabel{D} can be combined along the channel axis. In a partial execution loop, for each of the 144 channels, operator \boxlabel{B} (conv.) produces one output channel, operator \boxlabel{C} (depthwise conv.) transforms it, and operator \boxlabel{D} (conv.) consumes it and writes its contribution to its output buffer. The memory bottleneck is within the loop at operator \boxlabel{C}, amounting to 8.5 KB (at 8-bit precision). It consists of the input to \boxlabel{B}, one channel output of \boxlabel{B} (input to \boxlabel{C}), one channel output of \boxlabel{C}, and the full output buffer of \boxlabel{D}, preallocated before the start of the loop.

\begin{figure}[t]
    \centering
    \includegraphics[width=\linewidth]{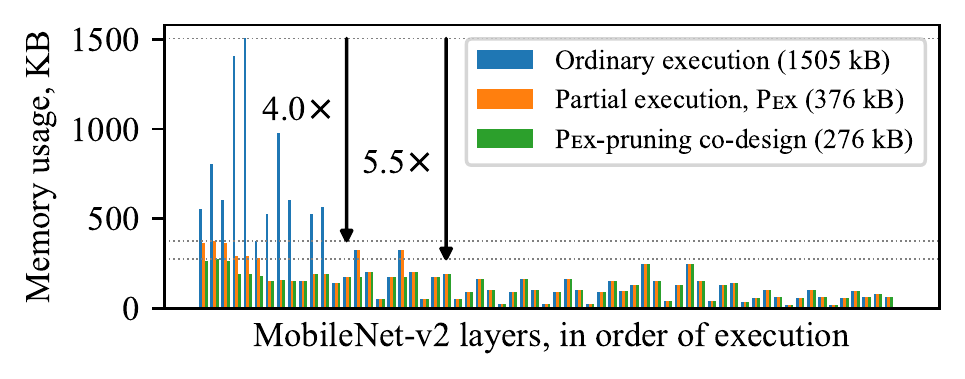}
    \vspace{-1em}
    \caption{Per-layer memory usage of MobileNet-v2. \textsc{Pex} automatically creates partial execution schedules that reduce peak memory usage of inference on microcontroller platforms. Memory is reclaimed at no extra computational cost, and the architecture can be pruned to reduce the memory bottleneck identified by \textsc{Pex}. Here, memory usage is reduced by 4$\times$, or 5.5$\times$ with pruning co-design.}
    \label{fig:mobilenetv2-pexec-pmu}
\end{figure}

Such execution has largely become obsolete on mobile platforms due to the abundance of RAM (Table~\ref{tab:devices-comp}) and an inferior data access pattern. However, on MCUs  data access pattern is of little concern due to the lack of data caches. Neural networks inspired by the IRB continue to proliferate~\cite{howard2019searching, cai2018proxylessnas, tan2019efficientnet}, so discovering analogous memory-saving execution schedules for new architectures would greatly benefit deep learning for MCUs by reducing the SRAM footprint of inference.

\textbf{Automatic partial execution with \textsc{Pex}.} We devise abstractions and execution rules required to apply partial execution to \emph{any} network layer layout, and \emph{automate} the scheduling decisions using a dynamic programming-based algorithm. These form the core of \textsc{Pex}, which discovers \emph{computationally-equivalent, yet more memory-efficient network execution schedules for arbitrary neural networks}. {\correction \textsc{Pex} produces an intermediate representation---a topologically-ordered list of network operators, their execution rules and partial execution loop information---which can be used to generate native microcontroller code.}

Figure~\ref{fig:mobilenetv2-pexec-pmu} shows the per-layer memory usage profile of the MobileNet-v2 architecture, showing that partial execution enables a \emph{4$\times$ peak memory usage reduction}. However, partial execution alone cannot reduce peak memory usage to arbitrarily low values: a particular layer arrangement at the memory bottleneck may not lend itself to optimisation, or the sizes of buffers allocated at the start of the partial execution loop may cause the memory bottleneck.

\textbf{Compiler+pruning network co-design.} We propose using \emph{structured pruning} to change the network architecture automatically to reduce peak memory usage under partial execution. Structured pruning entirely removes the least important feature maps (\emph{e.g.} convolutional channels), and we instruct it to target memory bottlenecks identified by the compiler. \textsc{Pex} is repeatedly invoked during pruning, with new memory bottlenecks reported back to the pruning algorithm at each step. Conversely, removing feature maps changes loop sizes and execution instructions produced by the compiler. Therefore, the two work in tandem to co-design the network architecture and the execution schedule.

We discuss the place of our methodology within related microcontroller deep learning work in Section~\ref{sec:related-work}, give the algorithm details in Section~\ref{sec:methodology-pexec}, including integration with pruning and considerations around quantisation, and perform quantitative comparisons in Section~\ref{sec:evaluation}.

Overall, the contributions of this work are as follows:
\begin{itemize}
\item We devise abstractions and execution rules for arbitrary networks, which enable partial execution.
\item We devise a model compiler, \textsc{Pex}, which automatically creates a partial execution schedule while minimising peak memory usage, unlocking significant memory usage reduction compared to ordinary evaluation.
\item We leverage partial peak memory usage as an optimisation goal within structured pruning, reducing the model’s memory footprint further automatically.
\item We reduce memory usage by up to 10.5$\times$ and establish new state-of-the-art models for three classification tasks in the low SRAM usage regime, improving accuracy by up to 2.9\%, compared to prior work.
\end{itemize}

\section{Related work}
\label{sec:related-work}
\subsection{Deep learning on microcontrollers}

Prior work in microcontroller deep learning typically assumes an ordinary execution-based runtime, such as TFLM \cite{david2020tensorflow}. MCU-friendly resource footprints are achieved by: manual architecture design~\cite{zhang2017hello, mocerino2019coopnet}, quantisation~\cite{jacob2018quantization}, network pruning~\cite{liberis2021differentiable} or neural architecture search (NAS)~\cite{lin2020mcunet, banbury2020micronets}. Some methods target inference with sparse matrices~\cite{fedorov2022udc, fedorov2019sparse}, or use pruning as a nested step within NAS~\cite{liberis2021munas}. Prior discovered models can operate within the partial execution regime.

Standardised runtimes are an important common ground for hardware, model compression and network design research. However, prior work shows that performance benefits can be reaped by moving beyond standard assumptions. TVM~\cite{chen2018tvm} is a general model compiler that can perform both computation graph- and inference code-level optimisations and produce binaries for various targets, including microcontroller platforms. MCUNet-v2~\cite{lin2021memory} forgoes layer-by-layer execution in favour of executing memory-intense layers using one input tile (“patch”) at a time; we discuss this execution style further on. 

Alternatively, computational constraints can be relaxed by introducing memory hierarchy. \citet{svoboda2022deep} use a microSD card to offload weights and activations of the network, which also significantly increases latency and power usage due to data transfer and the presence of external storage. More promisingly, \citet{sadiq2022tinyops} show that external SDRAM ($<$ 8 MB) and Flash memories can be used to offload selected parameters and activation matrices transparently using DMA. This approaches the latency of on-chip memory inference, but at the cost of additional power usage.

\subsection{Computation split approaches}
\label{sec:related-work-computation-split}

The computation of neural network layers’ outputs can also be split along their spatial axes instead of (or in addition to) the channel axis, resulting in \emph{patch-based execution}. 

Computation split approaches for memory-constrained microcontroller platforms overlap, to some degree, with the techniques developed for model parallelism. Conceptually, the two optimise for different scales: model parallelism maximises utilisation, cache hits, or reduces the multi-gigabyte memory (vRAM) footprint or communication overhead on multi-core GPU/TPU hardware for neural network training; here, we split computation to save 100s KB of SRAM for inference on relatively simplistic single-core hardware, but in the presence of quantisation and greater sensitivity to computation overheads (strict compute budget). 

The problems can be approached similarly: both develop frameworks to reason about neural networks, which identify independent computations and optimise their execution schedule. \citet{dryden2019improving} parallelise execution over the sample and spatial dimensions (\emph{i.e.} patches) and optimise workload assignment under a benchmark-informed cost model using a shortest-path graph algorithm; this methodology is also updated to consider channel and filter dimensions of a CNN~\cite{dryden2019channel}; {\correction Automap~\cite{schaarschmidt2021automap} and FlexFlow~\cite{jia2019beyond} automatically discover computation sharding axes by leveraging MCTS and predictor-powered randomised search, respectively;} \citet{xu2022training} combine patch-based execution with gradient checkpointing to reduce vRAM usage for high-resolution inputs; {\correction \citet{artemev2022xla} consider operation pattern matching and replacement, reordering and splitting of operators to conserve memory by reducing the size of intermediate matrices;}
 \citet{ivanov2021data} analyse data flow to  minimise data movement in Transformer models by identifying data layout and operator fusion improvements. We are optimistic about any future methodology cross-over.

In deep learning for MCUs, patch-based execution has been explored by MCUNet-v2~\cite{lin2021memory}. An analogous idea has been previously explored by \citet{alwani2016fused} for avoiding off-chip memory access. Only a spatial segment (a ``patch'') of a layer's output is computed at a time (thus full output is never materialised in memory), before being consumed by subsequent layers. This reduces memory usage at the cost of computation overhead due to the re-computing of intermediate values that are required for more than one patch (edge overlap). The amount of overhead varies per architecture and depends on convolutional kernel size and strides. Extended discussion of this alternative/orthogonal approach is in Appendix~\ref{apx:qualitative-patch-comparison}. As it is the closest relevant prior work, we perform a quantitative comparison in Section~\ref{sec:evaluation}.

\section{Design of Pex}
\label{sec:methodology-pexec}

\textsc{Pex} is a compiler for arbitrary neural networks which automatically generates execution schedules with partial execution loops (\emph{e.g.} Figure~\ref{fig:mobilenetv2-pexec-pmu}). This drastically reduces SRAM usage compared to ordinary execution used in prior work.

Conceptually, \textsc{Pex} is built as a progression of the following three components: (1) the partial execution framework: rules, axioms and abstractions, which define how each operator may be executed; (2) execution state and memory model, which allows deriving correct execution schedules for the entire network; (3) a dynamic-programming-based algorithm which finds an execution schedule that minimises the peak memory usage.

In the following, we describe the three components of $\textsc{Pex}$ and discuss the interaction between computational correctness and quantization, as well as using $\textsc{Pex}$ within structured pruning to reduce SRAM usage further. For contextual clarity, we will use convolutional neural network (CNN) terminology and reference the MobileNet-v2 IRB example.

\subsection{Partial execution framework}
\label{sec:pexec-rules}

\textbf{Definition 1.} A neural network is a computation graph of tensors, produced by operators (layers). We consider arbitrary neural networks which process $R$-dimensional tensors ($R \geq 2$) of shape $N \times \ldots \times C$, where $N$ is the batch size, and $C$ is the number of channels/feature maps. A "channel" is a tensor slice in the $C$ dimension. This includes 4-dimensional tensors of shape $N \times H \times W \times C$ seen in 2D CNNs, where $H$, $W$ are the width and height of the image. The dimensions' order is unimportant: for CNNs, NCHW and NHWC are supported; NCHW may be preferred to avoid strided loads/writes. 

\textbf{Definition 2.} A tensor $t$ can be present in memory in two forms: fully materialised (denoted by $t^F$) or partially materialised (only one channel, denoted by $t^P$). 

\textbf{Definition 3.} There are only two types of operators: channel\--wise operators (borrowing CNN terminology; denoted by `\emph{C}') and aggregating operators (denoted by `\emph{A}'):
\begin{itemize}
\item \textit{Channel-wise operators (C-types)} are depthwise convolutions, pooling layers (both local and global) and any element-wise operators such as addition or ReLU. In principle, these operators consume one input channel to produce one output channel. Output channels are independent, and the implementation of the operator is often a simple loop repeating the computation that maps an input channel to an output channel.
\item \textit{Aggregating operators (A-types)} are dot products: convolutions and fully connected (dense) layers. To compute the output tensor, they map all input channels to one output channel (multiple times, for all output channels), or, equivalently, compute one input channel’s contribution to all output channels (multiple times, for all input channels). That is, A-types can either generate one channel of their output at a time, providing the full input buffer is present in memory, or consume one channel at a time and write its contribution to the fully allocated output buffer. Due to this generating/aggregating property, they will be found either at the start or the end of partial execution loops.
\end{itemize}

\textbf{Remark 4.} We describe partial execution loops as processing one channel of data at a time. However, in a loop with a total of $C$ channels to be iterated over, the number of channels computed at once can be any $K \ll C$. Increasing $K$ increases the memory usage but can have a beneficial effect due to the use of SIMD kernel implementations (when $K$ is a multiple of 4) and lowering overhead associated with operator context switching, if any (implementation-dependent). The highest possible value of $K$ that does not cause a memory constraint violation can be found for each loop independently using binary search.

\textbf{Definition 5.} An operator is a mapping from input tensors $i=\{i_1, \ldots\}$ to an output tensor $o$, denoted $i \opexec o$. The operator's type (A or C) allows for multiple ways of executing it using different materialisation forms of inputs $i$ and output $o$. For operators with multiple inputs (\emph{e.g.} addition), all inputs always share the same materialisation form, so the set of inputs can be denoted with $i^F$ or $i^P$.

\begin{figure}[t]
    \centering
    \includegraphics[width=\linewidth]{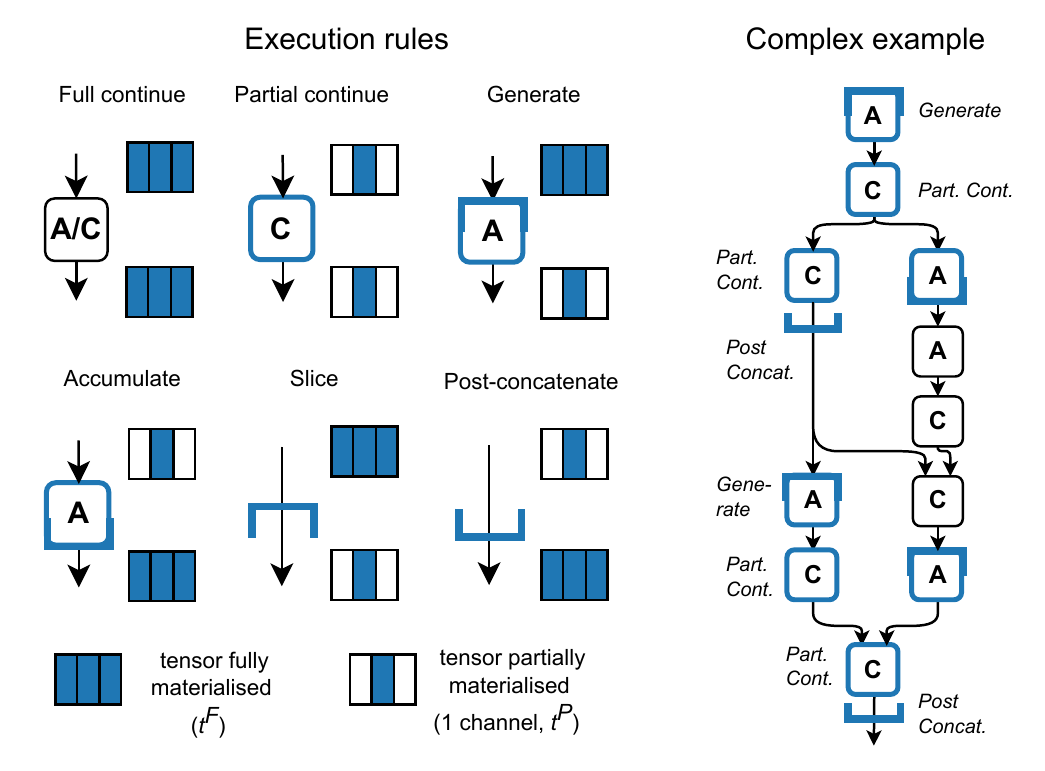}%
    \caption{\correction (\textit{Left:}) Diagrammatic representation of all partial execution rules. (\textit{Right:}) A complex computation graph example with multiple loops. Execution transitions between partially- and fully-materialised tensors at loop boundaries.}
    \label{fig:pex-rules-example}
\end{figure}

In total, we establish six rules for executing an operator $op$ (presented diagrammatically in Figure~\ref{fig:pex-rules-example}):

\begin{description}
\item[Full continue] (for both A- and C-types): $i^F \opexec o^F$. Input and output buffers are present in memory fully, and the operator is not in a partial execution loop. The compiler may fall back to ordinary execution by only using ‘Full Continue’. 
\item[Partial continue] (C-type only): $i^P \opexec o^P$. A channel-wise operator is executed for one input channel to produce one output channel. Example: a depthwise conv. operator computing one output channel.
\item[Generate] (A-type only). $i^F \opexec o^P$. Computes one output channel from fully-materialised input data. 
\item[Accumulate] (A-type only). $i^P \opexec o^F$ Consumes one channel of input data and writes its contribution to a fully-materialised output buffer. 
\end{description}

The following rules {\correction can be applied ``spontaneously`` to change the materialisation form of a tensor $t$, when needed.} For tensors consumed by multiple operators, the rule does not have to be applied to all branches (graph edges). 
\begin{description}
\item[Slice:] $t^F \longrightarrow t^P$. The runtime reads one channel at a time from tensor $t$ to facilitate partial execution of any subsequent operator that requires partially-materialised input. In this case, $t^P$ does not need to be explicitly allocated: it is an offset/view into $t^F$. 
\item[Post-concatenate:] $t^P \longrightarrow t^F$. Channels of $t$ are written into a preallocated full buffer. In this case, $t^P$ does not need to be explicitly allocated: the producer of $t^P$ can write into a particular offset of $t^F$.
\end{description}


\textbf{Definition 6.} A \emph{loop context} ($\mathcal{C}$) is a set of tensors that are kept in memory (fully materialised) due to a current loop that will repeatedly read from or write to these tensors. Loop context tensors are allocated in SRAM prior to the first iteration of the loop and deallocated after the last iteration.

In MobileNet-v2 IRB (Figure~\ref{fig:pex}), the context consists of inputs to operator \boxlabel{B} (exec. ‘Generate’) and the output of operator \boxlabel{D} (exec. ‘Accumulate’). At the $i^{th}$ iteration of the loop, \boxlabel{B} reads its entire input from the context to produce its $i^{th}$ channel, and \boxlabel{D} writes the contribution of its $i^{th}$ input channel to its output buffer in the context.

\begin{table*}[t]
\newcommand{\append}{\mkern-4mu :: \mkern-4mu}%

\caption{Execution state transition rules within \textsc{Pex}. The state is a tuple of tensors yet to be computed (`\texttt{Mem}') and loop context tensors (`$\mathcal{C}$'). Rules operate by decomposing `\texttt{Mem}' via pattern matching: a rule can be matched to any tensor of `\texttt{Mem}' via the left-hand side of the rule (providing conditions are satisfied; {\correction $A \opexec B$ conditions require the given transition to be valid within the network graph, executing an operator $op$, if needed}). The execution instructions (named after the state transition rules) are prepended (`::') to the list `$H$'.} 
\label{tab:pexec-rules}
\centering
\footnotesize
\begin{tabular}{l@{\hskip 5pt}r@{\hskip 2pt}c@{\hskip 2pt}l@{\hskip 6pt}l}
\toprule
\textbf{Rule} & \multicolumn{3}{c}{\correction \textbf{State transition}} & \textbf{Conditions} (Appendix~\ref{apx:derivation-rule-conditions}) \\
\midrule
End & $H \vdash \langle \texttt{Mem}, \varnothing \rangle$ & $\rightarrow$ & \textit{done} & $\langle \texttt{Mem}, \varnothing \rangle$ is $S_*$ (final state) \\
Loop End & $H \vdash \langle \texttt{Mem}, {\mathcal{C}} \rangle$ & $\rightarrow$ & $H \vdash \langle \texttt{Mem}, \varnothing \rangle$ & $\forall t \in \texttt{Mem}.$ $t$ is $t^F$ \\
Full Cont. & $H \vdash \langle \{o^F\} \cup R, {\mathcal{C}} \rangle$ & $\rightarrow$ & $\texttt{FullCont}(op) \append H \vdash \langle i^F \cup R, \varnothing \rangle$ & \boxlabel{$i^F \opexec o^F$}, \boxlabel{Eval.}, \boxlabel{No Loop} \\
Part. Cont. & $H \vdash \langle \{o^P\} \cup R, {\mathcal{C}} \rangle$ & $\rightarrow$ &  $\texttt{PartCont}(op) \append H \vdash \langle i^P \cup R, {\mathcal{C}} \rangle$ & \boxlabel{$i^P \opexec o^P$}, \boxlabel{Eval.}, \boxlabel{In Loop}, \boxlabel{C} \\
Generate & $H \vdash \langle \{o^P\} \cup R, {\mathcal{C}} \rangle$ & $\rightarrow$ & $ \texttt{Generate}(op) \append H \vdash \langle i^F \cup R, {\mathcal{C}} \cup i \rangle$ & \boxlabel{$i^F \opexec o^P$}, \boxlabel{Eval.}, \boxlabel{Comp. Loop}, \boxlabel{A}, $o \not\in {\mathcal{C}}$ \\
Aggregate & $H \vdash \langle \{o^F\} \cup R, \mathcal{C} \rangle$ & $\rightarrow$ & $\texttt{Aggregate}(op) \append H \vdash \langle i^P \cup R, {\mathcal{C}} \cup \{o\} \rangle$ & \boxlabel{$i^P \opexec o^F$}, \boxlabel{Eval.}, \boxlabel{Comp. Loop}, \boxlabel{A}, $o \not\in {\mathcal{C}}$ \\
Slice & $H \vdash \langle \{t^P\} \cup R, {\mathcal{C}} \rangle$ & $\rightarrow$ & $\texttt{Slice}(t) \append H \vdash \langle \{t^F\} \cup R, {\mathcal{C}} \cup \{t\} \rangle $ & \boxlabel{$t^F \longrightarrow t^P$}, \boxlabel{Comp.  Loop}, $t \not\in {\mathcal{C}}$ \\
Post-Concat. & $H \vdash \langle \{t^F\} \cup R, {\mathcal{C}} \rangle$ & $\rightarrow$ & $\texttt{PostConcat}(t) \append H \vdash \langle \{t^P\} \cup R, {\mathcal{C}} \cup \{t\} \rangle$ &
\boxlabel{$t^P \longrightarrow t^F$}, \boxlabel{Comp. Loop} \\
\bottomrule
\end{tabular}
\end{table*}

\subsection{Deriving an execution schedule}
\label{sec:execution-state}


A compiler produces a sequence of \emph{instructions}, which determine which operator should be executed (and how), to obtain the network's outputs from its inputs. Conceptually, an instruction is an action that causes the \emph{execution state} to transition from some current state to the next state. The instructions are applied until the final state is reached in which all neural network layers were executed.

We are interested in modelling memory usage, so the execution state will consist of information about tensors present in the memory, their materialisation form, and information about the current partial execution loop, if any.

\textbf{Definition 7.} The algorithm considers an execution state $S = \langle \texttt{Mem}, \mathcal{C} \rangle$ and builds up an instruction list (history) $H$. ‘\texttt{Mem}’ is the state of memory: a set of tensors currently present in memory, each either fully or partially materialised; ‘$\mathcal{C}$’ is the loop context (see Def. 6); $H$ is a list of instructions required to arrive at state $S$.

\textbf{Remark 8.} The network's parameters (weights) do not reside in SRAM under any execution regime---individual values are loaded from the read-only on-chip Flash memory.

\textbf{Definition 9.} The algorithm works backwards through the network computation graph $G$---tensors in `\texttt{Mem}' are treated as tensors yet to be computed. The initial state $S_0$ is the set of fully-materialised graph output tensors: $S_0 = \langle \{o^F | o \in \textnormal{outputs}(G)\}, \varnothing \rangle$ and, similarly, the final state $S_*$ is the set of fully-materialised inputs. This acts as a reachability analysis: only required operators will be executed. 

\textbf{Constraint 10.} No output is computed twice. We do not aim to increase the amount of computation required for a single inference pass of a neural network.

\textbf{Definition 11.} Table~\ref{tab:pexec-rules} lists allowed state transitions, called \emph{derivation rules}. Each rule has a set of conditions which must be satisfied for it to be applied. The transitions closely match the operator evaluation rules described earlier, but also modify the loop context as required. 

If derivation rules can be successfully applied to transition from the starting to the final state, we obtain a sequence of instructions (potentially, out of many) to execute a neural network. {\correction The instructions, along with any loop metadata, are sufficient for generating microcontroller code.} Figure~\ref{fig:pex} shows the MobileNet-v2 IRB example within the devised framework: first, by erasing particular layer information, leaving only A- and C-type annotations; then, by applying the derivation rules to produce a state transition graph. 

\subsection{An algorithm to minimise peak memory usage}
\label{sec:pexec-algorithm}

Out of numerous valid derivations, we are interested in choosing instructions that minimise SRAM usage. If we knew which sequence of instructions minimises peak memory usage from a particular execution state, we could record and reuse this information to build up a full execution schedule---a hallmark of a dynamic programming algorithm.

Algorithm~\ref{alg:optimise-pmu-func} describes a high-level pseudocode procedure `\texttt{OptimisePMU}' which enumerates all state transitions to find a sequence of instructions that results in the smallest peak memory usage (PMU) for arbitrary neural network computation graphs. Note that calls are memoised: \emph{results for the same input are remembered and not recomputed}. In practice, we are interested in not just accumulating execution history $H$ but also keeping track of the best-seen peak memory usage and the \emph{complete} loop context (denoted $\mathcal{C}^*$, returned from upstream computation) to calculate the PMU of operators within partial execution loops. The algorithm has exponential computational complexity in the number of operators but, due to small input size ($<$ 100 operators) and limited branching in the network architecture, the optimised schedules are obtained within seconds in practice.

\setlength{\textfloatsep}{10pt} 

\begin{algorithm}[t]
\caption{\textsc{OptimisePMU} computes a partial execution schedule which minimises peak memory usage.}
\label{alg:optimise-pmu-func}
\small
\algrenewcommand\algorithmicindent{1.0em}%
\begin{algorithmic}[1]
\State \Comment{\textit{Input:} a set of tensors to be computed \texttt{Mem} and loop context $\mathcal{C}$ (empty for the initial call).}
\State \Comment{\textit{Returns:} the instruction list $H$, the computed PMU and the final loop context $\mathcal{C}^*$ (empty iff $\mathcal{C} = \varnothing$).}

\Function{OptimisePMU}{\texttt{Mem}, $\mathcal{C}$}
\If{\texttt{End} (\texttt{Mem}, $\mathcal{C}$) \textit{conditions are satisfied}}
 \State \Return $\langle H{=}[], \textnormal{PMU}{=}\sum_{t \in \texttt{Mem}} |t|, \mathcal{C}^*{=}\varnothing \rangle$
\EndIf
\If{\texttt{LoopEnd} (\texttt{Mem}, $\mathcal{C}$) \textit{conditions are satisfied}}
 \State $H$, PMU, $\varnothing$ $\gets$ \Call{OptimisePMU}{\texttt{Mem}, $\varnothing$} 
 \State \Return $\langle H, \textnormal{PMU}, \mathcal{C} \rangle$
\EndIf

\State bestConfig $\gets$ \textit{(unknown)}

\For{$t \in \texttt{Mem}$}
 \For{\textit{Rule} $\in \{\texttt{FullContinue}, \texttt{Slice}, \ldots \}$}
  \If{\textit{Rule} (\texttt{Mem}, $\mathcal{C}$) \textit{conditions are satisfied}}
   \State $\texttt{Mem}'$, $\mathcal{C}'$ $\gets$ \textit{Rule}($t$, \texttt{Mem}, $\mathcal{C}$)
   \State $H'$, PMU$'$, $\mathcal{C}^*$ $\gets$ \Call{OptimisePMU}{$\texttt{Mem}'$, $\mathcal{C}'$}
   \State PMU$^*$ $\gets$ max(PMU$'$, \textsc{LocalPMU}($t$, $\texttt{Mem}'$, $\mathcal{C}^*$))
   \If{$\langle H', \textnormal{PMU}', \mathcal{C}^* \rangle$ \textit{better than} bestConfig}
    \State bestConfig $\gets$ $\langle H', \textnormal{PMU}', \mathcal{C}^* \rangle$
   \EndIf
  \EndIf 
 \EndFor
\EndFor
\State \Return bestConfig
\EndFunction
\end{algorithmic}
\end{algorithm}

The algorithm relies on four straightforward helper functions: (1) `\textit{rule conditions are satisfied}' checks if \texttt{Mem} and $\mathcal{C}$ satisfy set predicates set out in the definition of rule; (2) \textit{Rule(...)} invocation updates \texttt{Mem} and $\mathcal{C}$ according to the rule definition; (3) \textsc{LocalPMU}($t$, \texttt{Mem}, $\mathcal{C}^*$) computes the memory usage required (working set size) for tensor $t$ to be computed using tensors in \texttt{Mem}; if loop context is present, it will be counted towards the memory usage; (4) `\textit{configuration $x$ is better than $y$}' compares scheduling results: true if $x$ has lower peak memory usage, if $y$ is invalid, or if PMU is equal but $x$ is simpler; the latter can express a preference to avoid partial execution when there is no PMU improvement.


\setlength{\textfloatsep}{20pt} 

\subsection{Effects on quantisation}
\label{sec:quant-effects}

Neural network deployment on MCUs requires parameters and activations to be quantised. This both enables the execution and saves space: an MCU may not have floating-point processing units, or they may consume more power; Flash and SRAM usage is reduced when parameters and activations are stored at \emph{e.g.} 8-bit precision, instead of 32 bits.

Popular runtime and layer implementations use affine quantisation~\cite{jacob2018quantization}, where operators compute a single output element at a time (stored in a CPU register) by accumulating contributions from all input features (multiplied by the learned weight) at 32-bit precision. When all contributions have been added, an activation function is applied, and the output is written to SRAM at 8-bit precision. 

This poses a memory usage vs correctness trade-off for operators executed using the `Accumulate' partial execution rule, which accumulate input contributions at each loop iteration: the accumulation/output buffers must be kept in SRAM at 32-bit precision for computational correctness. This would result in a 4$\times$ memory usage increase for these buffers (changing from 8-bit to 32-bit representation), potentially reducing memory usage gains under partial execution.

To address this issue, we experiment with reducing the precision of the accumulation buffer, thus removing the 4$\times$ memory inflation. The ‘Accumulate’ rule would only apply to a few layers in the network; we expect other layers to learn to compensate for any performance hit caused by the reduced precision of affected convolutions. We experiment with quantization-aware training using 32-bit, 16-bit and 8-bit precision for accumulation buffers and find no performance degradation in the network (see Section~\ref{sec:quant-effects-analysis}).

\subsection{Structured pruning for partial execution}

Most deep learning models are too resource-demanding for MCU hardware, not just due to their memory usage but also size or latency, which can be addressed with model compression. Typically, three dimensions of resource usage are optimised for MCU compatibility: (a) peak memory usage (bounded by SRAM size), (b) model size (bounded by Flash memory) and (c) the number of multiply-accumulate operations (MACs, \citet{liberis2021munas} show that MACs are a highly accurate proxy for latency on an MCU processor).

We integrate \textsc{Pex} with the state-of-the-art MCU-aware model compression, which operates on the basis of differentiable structured pruning~\cite{liberis2021differentiable}, with the aim of (a) producing MCU-compatible models for our evaluation and (b) guiding pruning towards the memory bottleneck identified by our compiler. The method removes entire features/channels by repeatedly querying optimisation objectives during compression: we replace the default peak memory usage metric with an invocation of \textsc{Pex}, which reports the peak memory usage under partial execution and which operators' outputs constitute the bottleneck. 

\begin{table*}[t]
\caption{Resource usage of variations of MobileNet-v2 on ImageNet using ordinary, patch-based (4 $H$ and $W$ patches) or partial execution. PMU improvements brought by patch-based or partial execution vary, but partial execution always has zero computational overhead.}
\label{tab:mnetv2-patch-comparison}
\centering
\footnotesize
\begin{tabular}{lrrrrrrrrr}
\toprule
\multirow{2}{*}{\makecell[l]{\textbf{MobileNet-v2} \\ \textbf{variant}}} &
\multicolumn{2}{c}{\textbf{Loop overhead}} &
\multicolumn{2}{c}{\textbf{Total overhead}} &
\multicolumn{3}{c}{\textbf{Peak memory usage (PMU)}} &
\multirow[b]{2}{*}{\makecell[c]{\textbf{Accuracy} \\ \textbf{(8-bit q.)}}} &
\multicolumn{1}{c}{\textbf{MACs}} \\
\cmidrule{2-3} \cmidrule{4-5} \cmidrule{6-8} \cmidrule{10-10} 
& \textbf{Patch$^\dagger$} & \textbf{Partial} & \textbf{Patch$^\dagger$} & \textbf{Partial} & \textbf{Ordinary} & \textbf{Patch$^\dagger$} & \textbf{Partial} & & \textbf{Ordinary} \\
\midrule
Original & +30.6\% & \textbf{0\%} & +9.3\% & \textbf{0\%} & 1505 kB & 321 kB & 376 kB & 71.52\% & 301 M \\
RD (MCUNet-v2) & +13.8\% & \textbf{0\%} & +2.9\% & \textbf{0\%} & 1505 kB & 283 kB & 376 kB & 71.16\% & 294 M \\
Our modification & +38.9\% & \textbf{0\%} & +11.2\% & \textbf{0\%} & 978 kB & 278 kB & \textbf{226 kB} & 71.20\% & 295 M \\
\textsc{Pex}-pruning co-design & +36.1\% & \textbf{0\%} & +9.6\% & \textbf{0\%} & 1505 kB & 321 kB & 276 kB & 71.20\% & 288 M \\
Original, 172$\times$172 res. & +45.2\% & \textbf{0\%} & +12.6\% & \textbf{0\%} & 888 kB & \textbf{218 kB} & \textbf{222 kB} & 69.58\% & 193 M \\
\bottomrule
\end{tabular}
\end{table*}

This leads to \emph{co-design} between the model compression and the partial execution compiler. Both structured pruning and the compiler operate on the channel dimension of the neural network operators. Iteratively, the following takes place: (1) pruning adjusts per-layer channel counts based on the current memory bottleneck, and other resource usage objectives; (2) the compiler computes a new execution schedule using the updated operator channel dimensions: this new schedule is likely different due to new channel dimensions; (3) the compiler reports a new peak memory usage result and new memory bottleneck back to the pruning algorithm. 

\section{Evaluation}
\label{sec:evaluation}

In this section, we quantitatively evaluate our contributions:
\begin{enumerate}
    \item We compare partial execution with patch-based execution (MCUNet-v2), the closest relevant prior work, by considering peak memory usage reduction and overheads of either approach on MobileNet-v2 models.
    \item We apply \textsc{Pex} to five classes of architectures, using partial execution alone or the compiler-pruning co-design, to establish new state-of-the-art solutions in the low SRAM usage regime, compared to ten prior low-footprint models on three classification tasks.
    \item We test whether storing accumulation buffers at reduced precision affects models' accuracy (Section~\ref{sec:quant-effects}).
\end{enumerate}

\subsection{Comparison with patch-based execution}
\label{sec:patch-comparison}

Patch-based execution is an alternative approach of splitting the computation to reduce memory usage (Section~\ref{sec:related-work-computation-split}), which uses tensor slices along the image width and height axes (``patches''), instead of the channel axis. For MCUs, it is implemented by MCUNet-v2~\cite{lin2021memory}: we compare it to \textsc{Pex} by examining the memory usage reduction and computational overheads for the MobileNet-v2 model, followed by several modifications to it that enable further improvement in SRAM usage.

\renewcommand*{\thefootnote}{\fnsymbol{footnote}}

Table~\ref{tab:mnetv2-patch-comparison} (row 1) considers variations of MobileNet-v2 architecture, evaluated on ImageNet.\footnote[2]{The resource usage, overheads, and accuracy of MCUNet-v2 models have been adjusted from originally reported data due to differences in evaluation. We explain the rationale in Appendix~\ref{apx:quantitative-mcunet-adjustment}.}
The execution takes $\approx$301 M MACs and has a peak memory usage of 1505 kB. Patch-based and partial execution reduce it to 321 kB ($\downarrow$4.7$\times$) and 372 KB ($\downarrow$4$\times$), respectively, at the cost of additional 9.3\% increase in MAC operation count for patch-based execution yet \emph{zero computational overhead for our approach}.

\renewcommand*{\thefootnote}{\arabic{footnote}}

It is impossible to conclusively argue whether a patch-based or partial approach is generally superior at reducing peak memory usage: the outcome depends on the architecture. To illustrate this, we consider two modifications of MobileNet-v2 {\correction (see Figure~\ref{fig:mnetv2-mods} in Appendix~\ref{apx:qualitative-patch-comparison})}: (a) to reduce overheads of patch-based execution, MCUNet-v2 develop an alternative version with a redistributed receptive field (RD): a change to reduce the resolution and kernel size earlier in the network; (b) we present a version that reduces the memory bottleneck via additional spatial downsampling within the first inverted residual block (``MBConv''), in its projection layer, instead of the following block.
Table~\ref{tab:mnetv2-patch-comparison} (rows 2, 3) shows that these minor modifications can drastically affect the models’ resource usage: our variation of MobileNet-v2 significantly increases computation faced by patch-based execution compared to the original model (15.5\% total MAC increase) but, in the partial execution paradigm, yields the lowest peak memory usage seen so far with no computation overhead. 

The modifications above are specific to MobileNet-v2 and cannot be straightforwardly applied to other architectures. Therefore, we consider two other generic ways to reduce an architecture's peak memory usage: our compiler-pruning co-design and using lower input resolution. Table~\ref{tab:mnetv2-patch-comparison} (rows 4, 5) show that (a) co-design can reduce PMU under partial execution to match that of patch-based execution while offering comparable accuracy/MACs to the two previously considered modifications; (b) by lowering the input resolution, we observe the lowest peak memory usage achieved so far (218--222 KB) at the cost of a $<$ 2\% accuracy drop.

\textit{Summary of \ref{sec:patch-comparison}.} The network's architecture determines whether patch-based or partial execution will give greater memory reduction, and can be modified to benefit either approach. Partial execution has no computational overhead.

\subsection{Models under partial execution regime}
\label{sec:discovered-models}

In this section, we show the peak memory usage improvements achievable by \textsc{Pex} alone and the compiler-pruning co-design. We limit our scope to CNNs, as the most popular type of low-footprint neural networks, and consider three classification tasks: ImageNet (1000-class image classification)~\cite{deng2009imagenet}, VisualWakeWords (person detection)~\cite{chowdhery2019visual} and SpeechCommands (keyword recognition via spectrogram classification)~\cite{warden2018speech}, and five classes of architectures: MobileNet-v2~\cite{sandler2018mobilenetv2}, EfficientNet~\cite{tan2019efficientnet}, RES-15~\cite{tang2018deep}, MCUNet-v1 networks~\cite{lin2020mcunet} and MicroNets~\cite{banbury2020micronets}. Additional comparison points are provided by the differentiable pruning (DiffPru), DS-CNN~\cite{zhang2017hello}, MCUNet-v2~\cite{lin2021memory}, RES-8~\cite{tang2018deep} and ETinyNet~\cite{xu2022etinynet} architectures, resulting in ten low-footprint baselines. Models have been reevaluated under the same training pipeline and resource usage metrics (except for MCUNet-v2$^\dagger$). We use the peak memory usage (PMU) versus classification accuracy plots for comparison; tabular data with all metrics are available in Appendix~\ref{apx:architecture-tables}.

\begin{figure}[ht!]
    \centering
    \includegraphics[width=\linewidth]{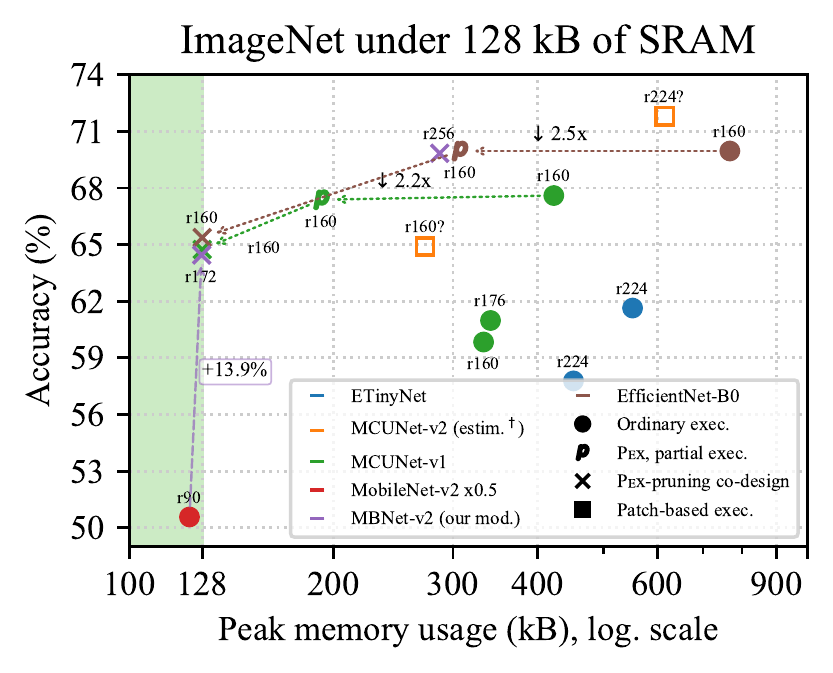}\\
    \includegraphics[width=\linewidth]{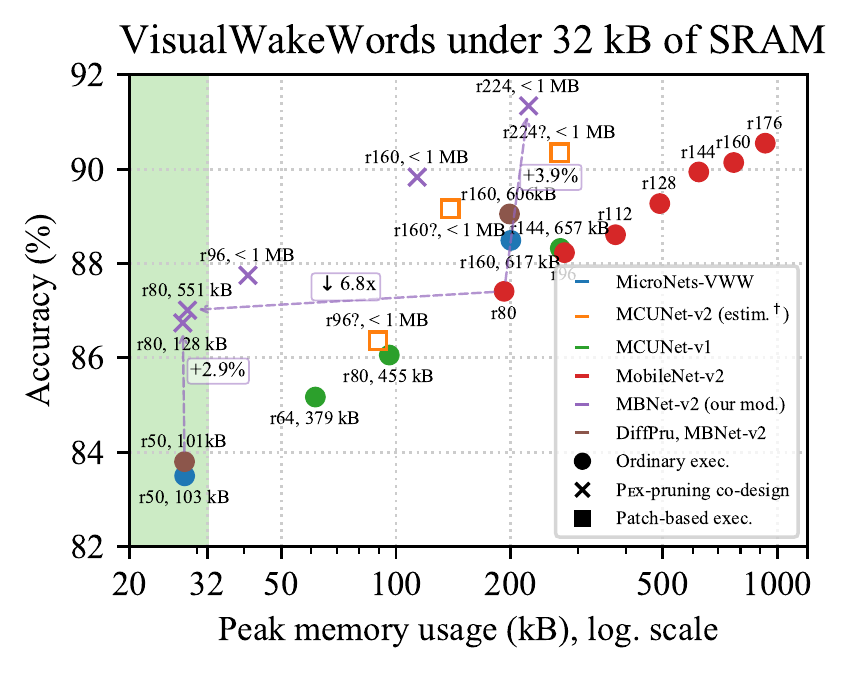}\\
    \includegraphics[width=\linewidth]{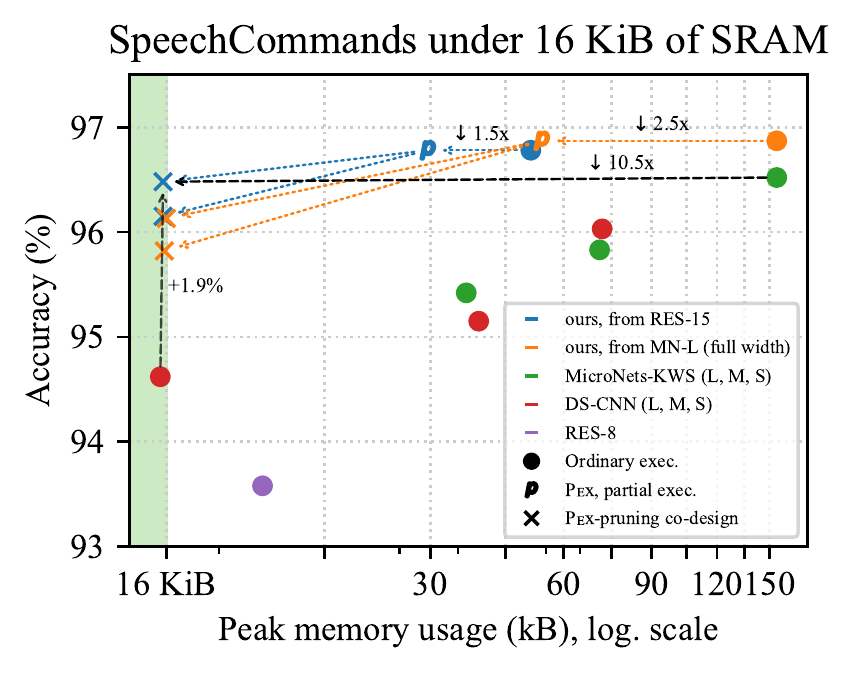}%
    \caption{Peak memory usage \emph{vs} accuracy on the considered datasets. The datapoint label, if present, denotes the input resolution (\emph{e.g.} ‘r160’ for 160x160 colour input) and, for MCU-compatible models, the size (storage requirement).}
    \label{fig:pareto-fronts}
\end{figure}

\textbf{ImageNet under 128 kB of SRAM.}
Figure~\ref{fig:pareto-fronts} (top) shows the peak memory usage versus accuracy trade-off for ImageNet models. \textsc{Pex} alone improves PMU by 2.2$\times$ and 2.5$\times$ for MCUNet-v1 and EfficientNet-B0 architectures. The compiler-pruning co-design produces \emph{the only set of high-performing models under 128 kB SRAM usage} amongst considered baselines in our evaluation environment. The models achieve 64.5\%--65.4\% accuracy, significantly outperforming a low-resolution MobileNet-v2 model (+13.9\%).

\textbf{VisualWakeWords under 32 kB of SRAM.} Figure~\ref{fig:pareto-fronts} (middle) shows the results for a variety of models for different input resolutions on the Visual Wake Words dataset. On MobileNet-v2, increasing input resolution increases PMU while producing diminishing returns in classification accuracy. At 80$\times$80 input resolution, the compiler-pruning co-design decreases the PMU by 6.8$\times$ compared to the baseline, bringing it within the 32 kB range. This constitutes a +2.9\% accuracy improvement compared to prior work (MicroNets VWW-2 and DiffPru) within this SRAM usage range. Alternatively, recovered SRAM allows using higher resolution inputs to \emph{increase accuracy by +3.9\% while retaining comparable SRAM usage} to the baseline.

\textbf{Keyword spotting under 16 KiB SRAM.} Figure~\ref{fig:pareto-fronts} (bottom) shows results for the SpeechCommands dataset. We apply both \textsc{Pex} alone, and the compiler-pruning co-design to the parent architecture of MicroNets-L (\emph{i.e.} instantiated at full channel counts, abbr. MN-L in the figure), and the RES-15 architecture. The use of compiler alone yields 2.5$\times$ and 1.5$\times$ memory reduction, respectively, and the use of pruning brings the models under the desired SRAM limit, with only up to 1\% accuracy loss compared to their original unpruned versions.  This constitutes a +1.9\% accuracy improvement compared to prior work (DS-CNN\--S) under the 16 KiB SRAM limit. Applying co-design to the RES-15 architecture yields \emph{up to 10.5$\times$ memory reduction while matching the classification accuracy} compared to the MicroNets-KWS-L baseline (using ordinary execution).

\textit{Summary of \ref{sec:discovered-models}.} \textsc{Pex} and \textsc{Pex}-pruning co-design reduce peak memory usage of high-performing models to establish state-of-the-art results in low SRAM usage regimes.

\subsection{Accumulation buffer precision analysis}
\label{sec:quant-effects-analysis}

Previously, in Section~\ref{sec:quant-effects}, we discussed how the interaction of quantisation and partial execution requires intermediate results of operators executed using the ‘Accumulate’ rule to be stored at 32-bit precision to preserve computational correctness.
To remedy the resulting memory usage increase, we suggest sacrificing computational correctness for affected operators by storing accumulation buffers at reduced precision and allowing other operators to learn to compensate for this imprecision. 

We aim to see whether reduced precision results in accuracy loss. We extended TensorFlow's quantisation-aware training (QAT) implementation to learn quantisation parameters for accumulation buffers (pre-activation), which are used during execution to re-quantise the buffers to the desired precision when updated. We experiment with the 32-bit (correct), 16-bit and 8-bit precision. The scheduler is parameterised with the memory cost of increased precision and, therefore, the ‘Accumulate’ rule will be avoided if there is an alternative with a lower PMU and fewer affected layers.

\begin{table}[t]
\caption{The impact of reduced accumulator precision for three diverse models. The data shows no to negligible accuracy loss from sacrificing computational correctness for PMU reduction.}
\label{tab:quant-accum-analysis}
\centering
\footnotesize
\begin{tabular}{lll}
\toprule
\makecell[c]{\textbf{MCUNet-v1-S} \\ \textbf{ImageNet 160x160}} &
\makecell[c]{\textbf{MobileNet-v2} \\ \textbf{VWW 160x160}} &
\makecell[c]{\textbf{MicroNets-KWS-L} \\ \textbf{SpeechCommands}} \\
\midrule
\multicolumn{3}{c}{\underline{\textbf{Mode:} Ordinary execution in \texttt{int8}}} \\[0.5em]
\makecell[l]{
PMU: 333 kB \\
Acc.: 59.89\% \\
0 accum. layers
} &
\makecell[l]{
PMU: 768 kB \\
Acc: 90.14\% \\
0 accum. layers
} &
\makecell[l]{
PMU: 170 kB \\
Acc: 96.77\% \\
0 accum. layers
} \\
\midrule
\multicolumn{3}{c}{\underline{\textbf{Mode:} Partial execution w/ 32-bit accumulators (QAT)}} \\[0.5em]
\makecell[l]{
186 kB ($\downarrow$1.8$\times$) \\
Acc.: 59.91\% \\
1 accum. layer
} &
\makecell[l]{
307 kB ($\downarrow$2.5$\times$) \\
Acc.: 90.14\% \\
0 accum. layers
} &
\makecell[l]{
69.0 kB ($\downarrow$2.5$\times$) \\
Acc.: 96.77\% \\
0 accum. layers
} \\
\midrule
\multicolumn{3}{c}{\underline{\textbf{Mode:} Partial execution w/ 16-bit accumulators (QAT)}} \\[0.5em]
\makecell[l]{
186 kB ($\downarrow$1.8$\times$) \\
Acc: 59.85 \\
1 accum. layer
} &
\makecell[l]{
294 KB ($\downarrow$2.6$\times$) \\
Acc: 90.12\% \\
2 accum. layers
} &
\makecell[l]{
69.0 kB ($\downarrow$2.5$\times$) \\
Acc: 96.71\% \\
1 accum. layer
} \\
\midrule
\multicolumn{3}{c}{\underline{\textbf{Mode:} Partial execution w/ 8-bit accumulators (QAT)}} \\[0.5em]
\makecell[l]{
141 kB ($\downarrow$2.4$\times$) \\
59.66\% (-0.23\%) \\
4 accum. layers
} &
\makecell[l]{
192 kB ($\downarrow$4$\times$) \\
Acc: 90.14\% \\
3 accum. layers
} &
\makecell[l]{ 
65.8 kB ($\downarrow$2.6$\times$) \\
Acc: 96.99\% \\
2 accum. layers
} \\
\bottomrule
\end{tabular}
\end{table}

Partial execution is mathematically equivalent to ordinary execution: without quantisation, both approaches have tensor values within the noise introduced by a differing order of operations; with quantisation, this noise is amplified, resulting in accuracy differences. We repeat the QAT for different quantisation levels of accumulation buffers; this also introduces noise due to randomness in the training pipeline, and the reduced precision also acts as a regulariser during QAT.

Table~\ref{tab:quant-accum-analysis} shows the classification accuracy of MCUNet-v1-S, MobileNet-v2, and MicroNets-KWS-L models under ordinary and partial execution, which shows no accuracy loss is observed for all tasks for up to 16-bit accumulation precision, resulting in up to 2.6$\times$ PMU reduction. Lowering the precision to 8-bit results in up to 4$\times$ PMU reduction and -0.23\% accuracy loss only in the MCUNet-v1-S experiment.

\textit{Summary of \ref{sec:quant-effects-analysis}.} Operators executed using the `Accumulate' rule can use reduced accumulation precision for lower memory usage: this has no or negligible accuracy loss.

\subsection{Limitations and future work}
\textsc{Pex}'s ability to reduce memory usage is limited by the network's architecture (also applies to patch-based execution, see Appendix~\ref{apx:qualitative-patch-comparison}), which can be adjusted in an informed way to allow further memory improvement. Here, we leveraged pruning as a general way to do so; \textsc{Pex} could be integrated with a neural architecture search system (NAS) that considers different layer connectivity to avoid some unfavourable layer arrangements automatically.

Additionally, data from an on-device MCU deployment could be of interest as auxiliary evaluation. {\correction The classification accuracy results presented in this section have been obtained by faithfully simulating partial execution under the presence of quantisation on a desktop machine.} However, because partial evaluation yields deterministic memory gains, we would expect any on-device deployment to only confirm this. 

\section{Conclusions}

A limited amount of SRAM on microcontroller platforms significantly impedes the on-device execution of neural networks on embedded, personal and IoT devices. To address this, we created a partial execution framework for arbitrary neural networks within the \textsc{Pex} model compiler, which automatically finds execution schedules that minimise peak SRAM usage. This is achieved by identifying subgraphs of the network's layers whose execution can be split along the channel axis, resulting in a computationally-equivalent yet more memory-efficient schedule compared to ordinary execution, at no extra computational cost. We enhance \textsc{Pex} with a structured pruning method, resulting in a co-design between the network's architecture and its execution schedule, which reduces SRAM usage even further. We evaluate \textsc{Pex} on five architecture classes and establish state-of-the-art performance in low SRAM usage regimes for ImageNet, VisualWakeWords and SpeechCommands datasets with up to +2.9\% accuracy increase, compared to ten prior low-footprint models. We find that a 4$\times$ memory usage reduction is possible by applying \textsc{Pex} alone, or up to 10.5$\times$ when using the compiler-pruning co-design, while maintaining the same classification accuracy. Alternatively, the recovered SRAM can be used to process higher resolution inputs, boosting accuracy by up to +3.9\% on Visual Wake Words.



\bibliography{paper}
\bibliographystyle{mlsys2023}

\appendix

\clearpage
\newpage

\section{Derivation rules}
\label{apx:derivation-rule-conditions}

\subsection{Rule conditions}

Each derivation rule listed in Table~\ref{tab:pexec-rules} has a number of conditions that need to be satisfied before the rule can be applied. Some conditions arise naturally from the pattern matching in the definition of the rule, \emph{e.g.} `Partial Continue' can only be applied to partially-materialised tensors in \texttt{Mem}.

\begin{description}
{\correction \item[\boxlabel{$A \opexec B$}] requires the computation from tensors $A$ to $B$ (in a certain materialisation form) to be a valid transition in the network computation graph. If present, an operator $op$ is executed as a part of this transition.}
\item[\boxlabel{In Loop}, \boxlabel{No Loop}] check if there is (or, respectively, is not) a partial execution loop currently being built ($C \not = \varnothing$). The presence of partially-materialised tensors in \texttt{Mem} implies \boxlabel{In Loop}.
\item[\boxlabel{A}, \boxlabel{C}] require the producing operator (`$op$') of the considered tensor to be of the aggregating or channel-wise type, respectively.
\item[\boxlabel{Eval.}] (can be evaluated) checks if an operator `$op$' with output $t$ can be executed within the current state. This is the case when $t$ is not a predecessor of (that is, $t$ is not required to compute) any other tensor in \texttt{Mem} (including itself in a different materialisation form). If ‘$t$’ were required to compute another tensor $t'$, $op$ would have to be executed again in the computation path for $t'$, resulting in redundant computation.
\item[\boxlabel{Comp. loop}] (compatible loop) checks if the operator can be added to the current partial execution loop (trivially true for 'Aggregate' and 'Post-Concatenate', if there is no current loop) by checking if its inputs' or output's (as appropriate for the rule) channel dimensions  match those of the current loop context tensors.
\item $t \not\in C$ checks if the tensor is not already in the loop context; used to remove trivial computation sequences (\emph{e.g.} 'Generate' followed by 'Post-Concatenate'). 
\end{description}

{\correction
\subsection{Rule intuition}
\textbf{Loop representation.} The proposed scheduling algorithm evolves the memory state $\langle \texttt{Mem}, {\mathcal{C}} \rangle$ and attempts to create partial execution loops by introducing partially-materialised tensors $t^P$ into $\texttt{Mem}$. The information about the current loop, if any, is represented by its $t^P$s in $\texttt{Mem}$. The loop context $\mathcal{C}$ is only used to store tensors of operators at the start or the end of the loop---those are tensors being repeatedly read from or written to, as loop execution proceeds. The context is completed and ``flushed'' (see ``Loop End'' rule) when the loop is finished (no $t^P$s remaining), and this complete context, returned through backtracking, is used to correctly measure the peak memory usage in the \textsc{LocalPMU} helper function in Algorithm~\ref{alg:optimise-pmu-func}.

\textbf{Loop correctness.} The rule conditions ensure that only one loop is being built at a time (\boxlabel{Comp. loop}), the loop building is not interrupted (\boxlabel{In/No loop}), and the correctness of the underlying network computation graph ensures matching loop iteration dimensions as the scheduling algorithm attempts to continue the loop by applying ``Partial Continue``, ``Generate`` and other rules.

\textbf{Loop fusion.} The scheduling can be alternatively cast a \emph{loop fusion} problem. Code snippets for each operator can be transformed to perform channel computation at the outermost level and the scheduler must determine the (topologically-correct) order of such snippets, as well as which outermost loops can be fused, to minimise the peak memory usage. 
In the presented framework, the choice of whether to continue advancing the loop (producing $i^P$) \emph{vs} closing it (producing $i^F$) and starting a new loop can be seen as a loop fusion decision.

\textbf{Facilitating code generation.} The primary purpose of the scheduling algorithm is to produce a list of instructions $H$ that describe the sequence of how each operator should be executed (which rule) in a topologically-correct order. To facilitate code generation, a variety of auxiliary data is captured by the algorithm, such as the loop dimensionality and the complete loop context for each instruction. These two properties are shared by all instructions belonging to the same loop, which, by definition, must form a continuous sequence within $H$. Upon encountering the first instruction of the loop, memory for loop context tensors is allocated (if not already present), and, similarly, context tensor memory is deallocated after the last introduction of the loop for any tensors not used later on.
}

\section{Qualitative comparison against patch-based execution}
\label{apx:qualitative-patch-comparison}

\begin{figure}[t]
    \centering
    \includegraphics[width=\linewidth]{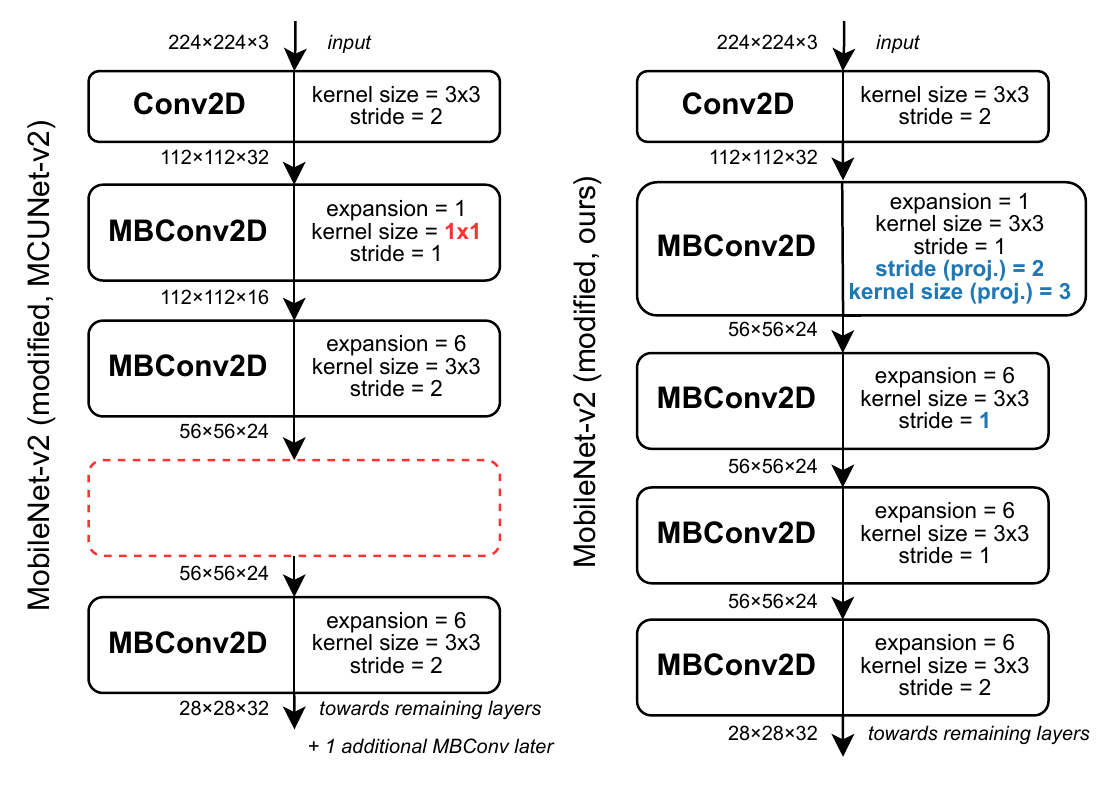}%
    \caption{Modified versions of MobileNet-v2 which, compared to the original architecture, allow greater SRAM usage reduction within patch-based (\textit{left}) or partial (\textit{right}) execution (Table~\ref{tab:mnetv2-patch-comparison}).}
    \label{fig:mnetv2-mods}
\end{figure}

Instead of the channel axis $C$, the computation can also be split along spatial axes $H, W$. In MCU deep learning, patch-based execution is exemplified by MCUNet-v2~\cite{lin2021memory}, and in the following, we discuss the advantages, disadvantages and trade-offs of the two approaches.

{\correction 
\textbf{The performance is architecture-dependent.} Results presented in Section~\ref{sec:patch-comparison} established that the improvement in memory usage depends on the underlying architecture. That is, it is not possible to conclusively argue that partial or patch-based execution always yields a greater improvement. The considered modifications to the MobileNet-v2 architecture are presented in Figure~\ref{fig:mnetv2-mods}.
}

\textbf{Redundant computation in patch-based execution.} As patch-based computation transitions between adjacent output ``pixels'' of a convolutional operator, it may use the same spatial regions of the input (overlapping receptive field) to obtain the output value. If the input patch is deallocated during this transition, this shared fraction of the input will have to be recomputed, leading to computational overheads (we refer to \citet{lin2021memory} for a more detailed diagrammatic explanation of this issue). This can be partly alleviated by minimising overlapping input regions for patch-executed operators, which authors do both manually and automatically by searching and adjusting, among other parameters, strides, and kernel and patch sizes within the MobileNet-v2-based backbone. In contrast, our methodology does not result in any redundant computation (Constraint 5). 

\textbf{Both techniques suffer under unfavourable layer arrangements.} We showcase an unfavourable case for both approaches: squeeze-excitation (SE) layers in state-of-the-art architectures, such as MobileNet-v3 (small, large) and EfficientNet-B0, which prevent both methods from achieving greater peak memory reduction. 

\begin{figure}[t]
    \centering
    \includegraphics[width=\linewidth]{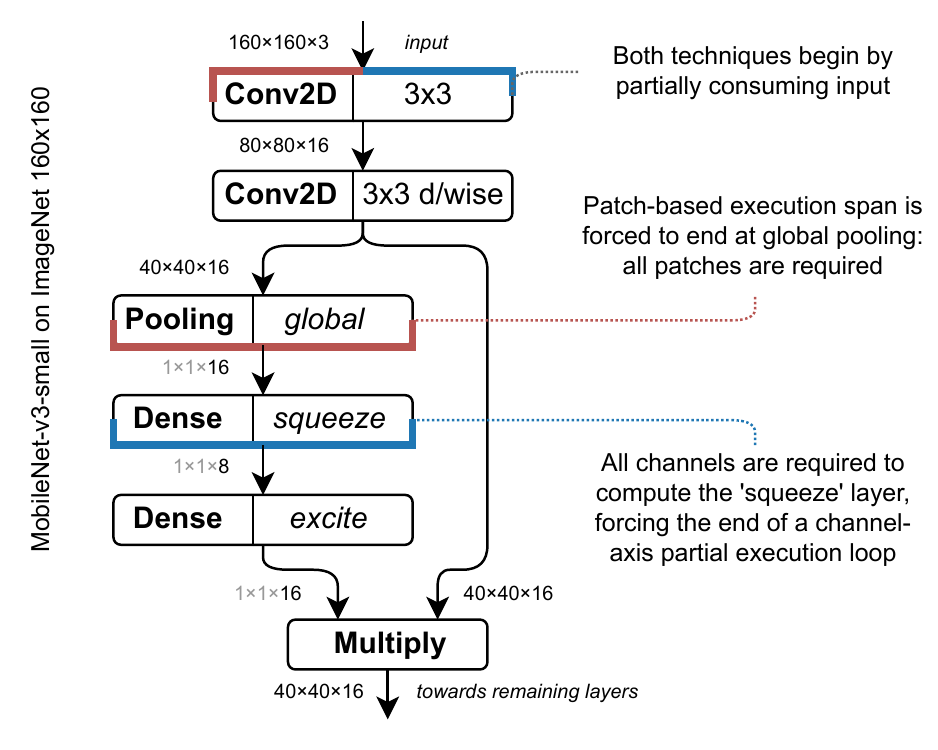}%
    \caption{Unfavourable layer arrangement of \emph{MobileNet-v3-small}.}
    \label{fig:mnetv3-unfavourable}
\end{figure}

Figure~\ref{fig:mnetv3-unfavourable} shows the first six layers of the MobileNet-v3-small architecture, which contain squeeze-excitation layers: a global pooling layer (reduction along all spatial dimensions), followed by compression and expansion of the channel dimension using two fully-connected layers (‘squeeze’ and ‘excite’). Reduction across all spatial dimensions requires values from all input patches, which ends the patch-based execution span; similarly, the subsequent ‘squeeze’ layer requires values from all input channels, which terminates the current partial execution loop. Of course, a new partial execution loop can be started further in the network.

A straightforward workaround is modifying the architecture by removing the squeeze-excitation bottleneck from the high SRAM usage parts of the architecture. For example, a sibling architecture, MobileNet-v3-large, does not contain SE layers within the first three inverted residual blocks.

\textbf{Greater memory reduction potential for very high spatial resolution low-channel layers with patch-based execution.} When $H, W \gg C$, patch-based execution offers a greater memory reduction potential: in the limit, $1/HW$ for a 1$\times$1 patch (and, in that instance, a significant re-computation overhead) versus $1/C$ for one channel. Therefore, some high-resolution computer vision neural networks, such as those processing aerial or medical imaging data, may benefit from patch-based execution more, depending on their architecture.

We also note that these approaches are not mutually exclusive: channel-axis partial execution can be applied within patch-based execution to reduce SRAM usage and enable the use of larger tiles, thus lowering the computation overhead. Both approaches can be unified within a more complex framework that jointly considers inter-operator tiling in spatial and channel dimensions.

The discussion above would suggest that channel-axis partial execution offers lesser peak memory usage gains at the benefit of no redundant computation and thus faster execution. We quantitatively examine this claim in Section~\ref{sec:evaluation} by evaluating both execution approaches on a high-resolution (for MCU applications) MobileNet-v2 model trained for ImageNet classification. Specifically, we:
\begin{enumerate}
    \item quantitatively evaluate differences in peak memory usage and computation overhead for both approaches; 
    \item examine how these quantities change with minor modifications to the architecture that would make it more favourable to either approach; 
    \item consider an automatically compressed model using structured pruning, instead; 
    \item check if the high-resolution input is actually necessary for the classification task at hand.
\end{enumerate}

\section{Adjustments for the quantitative comparison against MCUNet}
\label{apx:quantitative-mcunet-adjustment}

There are a number of adjustments that need to be made to the network performance and resource usage reported in the MCUNet-v1 and -v2 manuscripts for a fair evaluation.

\textbf{Peak memory usage (MCUNet-v2 only).} The numbers presented in Evaluation (Section~\ref{sec:evaluation} differ from the MCUNet-v2 manuscript, which departs from the evaluation settings of prior work, such as MCUNet-v1~\cite{lin2020mcunet}, differentiable pruning~\cite{liberis2021differentiable} or MicroNets~\cite{banbury2020micronets}, by assuming that the full input tensor does not need to be stored in memory. Instead, compressed JPEG-encoded input ought to be available in memory for partial decoding. However, this assumption does not generally hold: while true for applications that have hardware with JPEG camera modules, it would not be the case for audio or sensor data processing applications. Compression of inputs and/or feature maps (and associated additional power usage and decoding overheads) is an interesting but, ultimately, tangential research direction to evaluating execution techniques, so we recompute patch-based execution memory usage within the same general environment for a fair comparison. For cases where the network architecture definition is not available, we assume that peak memory usage occurs in the patch-based execution stage and add the input size to the reported memory usage.

\textbf{Classification accuracy on Visual Wake Words (VWW) (-v1 and -v2).} The authors of VWW dataset provide training and validation data splits and suggest using the ``minival'' dataset (a small subset of the validation split) for final reporting. However, the ``minival'' happens to be a biased sample of the validation set, which results in noisier and inflated classification accuracy: models claiming to achieve $>$94\% accuracy only achieve $\approx$90\% on a larger validation set. Some prior work~\cite{liberis2021differentiable, banbury2020micronets, saha2020rnnpool} uses the full validation set for evaluation, instead of ``minival''. For MCUNet-v1, architecture definitions are available, allowing us to re-evaluate the models (and observe the accuracy disparity between the two evaluation sets). For MCUNet-v2, definitions are unavailable; thus we linearly extrapolate from MCUNet-v1 results, assuming both use the same training pipeline.

\textbf{Miscellaneous differences.}
\begin{itemize}
    \item We note different classification accuracies for the ImageNet dataset (MobileNet-v2 performing at 72.2\% in the MCUNet-v2 manuscript \emph{vs} ours’ 71.5\%) due to differences in the training pipeline. As these differences are under 1\%, we consider them to be negligible. 
    \item When the input resolution is not reported, we extrapolate from MobileNet-v2 results as to what input resolution would be appropriate to achieve reported resource usage or classification accuracy.
\end{itemize}

We acknowledge that differing evaluation assumptions and adjustments make the comparison challenging. Still, in the absence of other related work, we believe that the data reported here paint a fairer picture of the relative performance of partial and patch-based execution on common tasks.

\section{Architecture resource usage}
\label{apx:architecture-tables}

The following tables list detailed resource usage and classification accuracy data of models presented in Figure~\ref{fig:pareto-fronts} in Section~\ref{sec:evaluation}. As with the Figure, icons next model name represent ordinary execution ($\bullet$), partial execution ($p$), pruning-compiler co-design ($\times$) or patch-based execution ($\square$). MCUNet-v2 results are adjusted and estimated ($\dagger$): see Appendix~\ref{apx:quantitative-mcunet-adjustment}.

\newpage
\centering
\footnotesize

Properties of ImageNet models.
\begin{tabular}{lr@{\hskip 9pt}r@{\hskip 9pt}r@{\hskip 9pt}r}
\toprule
\textbf{Model} & \textbf{PMU} & \textbf{Size} & \textbf{MACs} & \textbf{Acc.} \\
\midrule
ETinyNet x1.00$^\bullet$ & 552 kB & 979 kB & 114 M & 61.7\% \\
ETinyNet x0.75$^\bullet$ & 452 kB & 660 kB & 69.7 M & 57.8\% \\
MCUNet-v2 M4$^{\square,\dagger}$ & 273 kB & 1010 kB & 119 M & 64.9\% \\
MCUNet-v2 H7$^{\square,\dagger}$ & 616 kB & 2032 kB & 256 M & 71.8\% \\
MB-v2 (mod.) r172$^\times$ & 128 kB & 999 kB & 110 M & 64.5\% \\
MB-v2 (mod.) r256$^\times$ & 287 kB & 1994 kB & 255 M & 69.8\% \\
MCUNet-v1-S$^\bullet$ & 333 kB & 748 kB & 67.4 M & 59.8\% \\
MCUNet-v1-M$^\bullet$ & 341 kB & 756 kB & 81.9 M & 61.0\% \\
MCUNet-v1-L$^\bullet$ & 422 kB & 1757 kB & 126 M & 67.4\% \\
MCUNet-v1-L$^p$ & 192 kB & 1757 kB & 126 M & 67.3\% \\
MCUNet-v1-L$^\times$ & 128 kB & 994 kB & 88.9 M & 63.5\% \\
EfficientNet-B0$^\bullet$ & 768 kB & 3987 kB & 187 M & 69.2\% \\
EfficientNet-B0$^p$ & 308 kB & 3987 kB & 187 M & 69.2\% \\
EfficientNet-B0$^\times$ & 128 kB & 1999 kB&  118 M & 65.4\% \\
EfficientNet-B0$^\times$ & 128 kB & 1999 kB&  118 M & 65.4\% \\
MBNet-v2 x0.5 r90$^\bullet$ & 123 kB & 1990 kB & 18 M & 50.6\% \\
\bottomrule
\end{tabular}

Properties of Visual Wake Words models.
\begin{tabular}{l@{\hskip 8pt}r@{\hskip 8pt}r@{\hskip 8pt}r@{\hskip 6pt}r}
\toprule
\textbf{Model} & \textbf{PMU} & \textbf{Size} & \textbf{MACs} & \textbf{Acc.} \\
\midrule
MicroNets VWW-1$^\bullet$ & 200 kB & 616 kB & 71.6 M & 88.5\% \\
MicroNets VWW-2$^\bullet$ & 27.9 kB & 103 kB & 3.38 M & 83.5\% \\
MCUNet r144$^\bullet$ & 270 kB & 657 kB & 55.9 M & 88.3\% \\
MCUNet 5FPS$^\bullet$ & 96.0 kB & 455 kB& 12.5 M & 86.1\% \\
MCUNet 10FPS$^\bullet$ & 61.4 kB & 379 kB & 5.97 M & 85.2\% \\
MCUNet-v2$^{\square,\dagger}$ & 268 kB & \textit{unk.} & \textit{unk.} & $\approx$90.4\% \\
MCUNet-v2$^{\square,\dagger}$ & 139 kB & \textit{unk.} & \textit{unk.} & $\approx$89.2\% \\
MCUNet-v2$^{\square,\dagger}$ & 89.6 kB & \textit{unk.} & \textit{unk.} & $\approx$86.4\% \\
MB-v2 (mod.) r224$^\times$ & 223 kB & 1 MB & 209 M & 91.3\% \\
MB-v2 (mod.) r160$^\times$ & 114 kB & 1 MB & 106 M & 89.8\% \\
MB-v2 (mod.) r96$^\times$ & 40.8 kB & 1 MB & 37.7 M & 87.8\% \\
MB-v2 (mod.) r80$^\times$ & 28.4 kB & 511 kB & 21.2 M & 87.0\% \\
MB-v2 (mod.) r80$^\times$ & 27.6 kB & 128 kB & 11.8 M & 86.7\% \\
MB-v2 r80$^\bullet$ & 192 kB & 2.3 MB & 42.7 M & 87.4\% \\
MB-v2 r96$^\bullet$ & 276 kB & 2.3 MB & 55.1 M & 88.2\% \\
MB-v2 r112$^\bullet$ & 376 kB & 2.3 MB & 81.0 M & 88.6\% \\
MB-v2 r128$^\bullet$ & 492 kB & 2.3 MB & 97.9 M & 89.3\% \\
MB-v2 r144$^\bullet$ & 622 kB & 2.3 MB & 132 M & 89.9\% \\
MB-v2 r160$^\bullet$ & 768 kB & 2.3 MB & 153 M & 90.1\% \\
MB-v2 r176$^\bullet$ & 929 kB & 2.3 MB & 194 M & 90.6\% \\
DiffPru, MB-v2$^\bullet$ & 198 kB & 606 kB & 58.3 M & 89.1\% \\
DiffPru, MB-v2$^\bullet$ & 27.9 kB & 101 kB & 3.34 M & 83.8\% \\
\bottomrule
\end{tabular}

Properties of Speech Commands models.
\begin{tabular}{l@{\hskip 9pt}r@{\hskip 9pt}r@{\hskip 9pt}rr}
\toprule
\textbf{Model} & \textbf{PMU} & \textbf{Size} & \textbf{MACs} & \textbf{Acc.} \\
\midrule
MicroNets-KWS-L$^\bullet$ & 170 kB & 512 kB & 65.7 M & 96.5\% \\
MicroNets-KWS-M$^\bullet$ & 86.1 kB & 117 kB & 15.6 M & 95.8\% \\
MicroNets-KWS-S$^\bullet$ & 51.6 kB & 63.5 kB & 8.4 M & 95.4\% \\
RES-8$^\bullet$ & 23.7 kB & 112 kB & 4.16 M & 93.6\% \\
RES-15$^\bullet$ & 66.2 kB & 240 kB & 116 M & 96.8\% \\
RES-15$^p$ & 44.6 kB & 240 kB & 116 M & 96.8\% \\
RES-15$^\times$ & 16.1 kB & 86.7 kB & 41.4 M & 96.5\% \\
RES-15$^\times$ & 16.1 kB & 55.8 kB & 26.6 M & 96.2\% \\
MicroNets-L-Base$^\bullet$ & 170 kB & 582 kB & 74.3 M & 96.9\% \\
MicroNets-L-Base$^p$ & 69 kB & 582 kB & 74.3 M & 96.9\% \\
MicroNets-L-Base$^\times$ & 16.4 kB & 144 kB & 20.5 M & 96.1\% \\
MicroNets-L-Base$^\times$ & 16.3 kB & 88.6 kB & 12.5 M & 95.8\% \\
DS-CNN-L$^\bullet$ & 86.9 kB & 420 kB & 28.3 M & 96.0\% \\
DS-CNN-M$^\bullet$ & 54.2 kB & 140 kB & 9.83 M & 95.2\% \\
DS-CNN-S$^\bullet$ & 16.0 kB & 24.3 kB & 2.66 M & 94.6\% \\
\bottomrule
\end{tabular}


\end{document}